\documentclass[conference]{IEEEtran}
\IEEEoverridecommandlockouts
\usepackage{cite}
\usepackage{amsmath,amssymb,amsfonts}
\usepackage{algorithmic}
\usepackage{graphicx}
\usepackage{textcomp}
\usepackage{xcolor}
\usepackage[caption=false]{subfig}
\usepackage[hyphens]{url}

\usepackage{fancyhdr}
\begin{document}

\title{Go-Blend Behavior and Affect
\thanks{This project has received funding from the European Union’s Horizon 2020 programme under grant agreement No 951911.}
}

\author{
\IEEEauthorblockN{Matthew Barthet}
\IEEEauthorblockA{\textit{Institute of Digital Games} \\
\textit{University of Malta}\\
Msida, Malta \\
matthew.barthet@um.edu.mt}
\and
\IEEEauthorblockN{Antonios Liapis}
\IEEEauthorblockA{\textit{Institute of Digital Games} \\
\textit{University of Malta}\\
Msida, Malta \\
antonios.liapis@um.edu.mt}
\and
\IEEEauthorblockN{Georgios N. Yannakakis}
\IEEEauthorblockA{\textit{Institute of Digital Games} \\
\textit{University of Malta}\\
Msida, Malta \\
georgios.yannakakis@um.edu.mt}
}

\maketitle
\thispagestyle{fancy}

\begin{abstract}
This paper proposes a paradigm shift for affective computing by viewing the affect modeling task as a reinforcement learning process. According to our proposed framework the context (environment) and the actions of an agent define the common representation that interweaves behavior and affect. To realise this framework we build on recent advances in reinforcement learning and use a modified version of the Go-Explore algorithm which has showcased supreme performance in hard exploration tasks. In this initial study, we test our framework in an arcade game by training Go-Explore agents to both play optimally and attempt to mimic human demonstrations of arousal. We vary the degree of importance between optimal play and arousal imitation and create agents that can effectively display a palette of affect and behavioral patterns. Our Go-Explore implementation not only introduces a new paradigm for affect modeling; it empowers believable AI-based game testing by providing agents that can blend and express a multitude of behavioral and affective patterns.     
\end{abstract}


\IEEEpeerreviewmaketitle

\section{Introduction}\label{sec:introduction}

Affective computing is traditionally viewed from an expert-domain and supervised learning lens through which manifestations of affect are linked to ground truth labels of affect that are provided by humans. Behavior and affect are either blended in the form of hand-crafted rules \cite{marsella2010computational,marsella2009ema} or machine learned via supervised learning methods \cite{calvo2010affect}. While affect models designed or built this way are linked to the context of the interaction, they are often completely independent of the behavior of the involved actors. 

A recent (non-deep) reinforcement learning (RL) algorithm, Go-Explore \cite{ecoffet2018montezuma}, showcased superb performance at hard exploration problems with many states---such as complex planning-based games---that most other deep learning methods struggled with. In its application to the game \emph{Montezuma's Revenge} (Parker Brothers, 1984), Go-Explore reached super-human gameplaying performance. In part, this is achieved by storing all visited game states and exploring from such interim states rather than playing the game from the start \cite{ecoffet2021first}. Inspired by these recent breakthroughs in RL, we leverage the capacity of Go-explore to introduce a paradigm shift for affect modeling. We argue that viewing affect modeling as an RL process yields agents (or computational actors) that manage to reliably interweave behavior and affect without necessarily relying on affect corpora of massive sizes.

The proposed concept revolutionizes affective computing, which traditionally attempts to model human affect in the context of an interaction but largely ignores the affective response to the \emph{actions} of the involved (inter-)actors. Both behavior and affect are blended in an internalised model that associates an agent's context (environment) and its actions to both its behavioral performance and its affective state. At the same time, we introduce a novel paradigm for RL where the rewards are not only tied to a user's behavior but combined with rewards from annotations provided by the users themselves (i.e. human affect demonstrations). According to our approach, both behavior and affect can form reward functions that can be experienced from RL agents that learn to behave and express affect in various ways. The proposed Go-Explore implementation is tested in a simple arcade game featuring a rich corpus of self-reported traces of arousal. 

Our key findings suggest that agents can be trained effectively to behave in particular ways (e.g. play optimally with super-human performance) but also behave so as they \emph{feel} as humans would in a particular game state. Beyond the proposed paradigm shift in affective computing, our Go-Explore agents offer insights on the relationship between affect and behavior through their RL trained models. Importantly, RL agents that blend behavior and affect can be used directly for believable testing as such agents can simulate and express simultaneously both behavioral and affective patterns of humans. 

\section{Background}\label{sec:relatedwork}

This section provides a brief overview of the related domains of reinforcement learning, the Go-Explore algorithm, traditional affect modelling via imitation learning and affect modelling using reinforcement learning.

\subsection{Reinforcement Learning and Go-Explore}\label{sec:relatedwork_reinforcement}

Reinforcement learning approaches machine learning tasks from the perspective of behavioral psychology, mimicking the way animals and humans learn through receiving positive or negative rewards for their actions \cite{sutton2018reinforcement}. Exploring state spaces with sparse and/or deceptive rewards has been a core challenge for traditional RL algorithms, as they suffer from issues of detachment and derailment. \emph{Detachment} occurs when an algorithm forgets how to return to previously visited promising areas of the search space due to exploration in other areas. \emph{Derailment} is a consequence of RL algorithms which do not separate returning to states from exploring the search space. This may result in potentially promising states that require a long sequence of precise actions unlikely to occur under exploratory conditions. 

Go-Explore is a recent algorithm in the RL family \cite{ecoffet2019go} which is explicitly designed to overcome the two aforementioned challenges. The algorithm was introduced with the aim of improving RL performance in hard-exploration problems, which tend to contain sparse or deceptive rewards. Go-Explore has demonstrated previously unmatched performance in Atari games \cite{ecoffet2021first}, highlighting its ability to thoroughly explore complex and challenging environments. In games with sparse rewards (such as \emph{Montezuma's Revenge}), a large number of actions must be taken before a reward can be obtained, whereas deceptive rewards may mislead the agent and result in premature convergence and therefore poor performance \cite{yannakakis2018artificial}. Go-Explore has been used for text-based games, capable of outperforming traditional agents in \emph{Zork1} \cite{ammanabrolu2020avoid} and is able to generalize to unseen text-based games more effectively \cite{madotto2020exploration}. The algorithm's capabilities have also been demonstrated in complex maze navigation tasks which could not be completed by traditional RL agents \cite{matheron2020pbcs}. Beyond playing planning-based games with superhuman performance, Go-Explore has been used for autonomous vehicle control for adaptive stress testing \cite{koren2020adaptive}, and as a mixed-initiative tool for quality-assurance testing using automated exploration \cite{chang2019reveal}. 
While Go-Explore has proved to be a highly effective algorithm for behaviour policy search, it has never been tested on affect modeling tasks. This proof-of-concept paper introduces the first application of the algorithm for modeling affect as an RL process and blending it with behavior within a game agent. 

\subsection{Reinforcement Learning and Affective Computing}\label{sec:relatedwork_affective}

Traditionally, affect modelling \cite{calvo2010affect} involves constructing a computational model of affect that takes as input the context of the interaction, such as pixels \cite{makantasis2019pixels,makantasis2021pixels}, and multimodal information about a user---including physiological signals \cite{martinez2013learning}, facial expressions \cite{ruiz2018multiinstance,walecki2017deep} or speech \cite{trigeorgis2016adieu}---and outputs a predicted corresponding emotional state (i.e. the ground truth of emotion). Given that affective computing relies on a provided ground truth of emotion that is human-annotated, affect detection is naturally viewed as a supervised learning task \cite{calvo2010affect}. Traditionally a dataset of user state-affect pairs is used to train a model to predict affect \cite{Koelstra2012DEAPAD}. Trained affect models are then used in conjunction with action selection methods for the synthesis, adaptation and affect-based expression of agents including virtual humans \cite{swartout2006toward} and social believable agents \cite{reilly1996believable}. 

Beyond the obvious uses of RL for learning a behavior policy, RL has been used as a paradigm for creative AI and, in particular, for the procedural generation of content (PCG) \cite{khalifa2020pcgrl}. While the experience-driven PCG framework \cite{yannakakis2015experience} considered the use of affect models beyond the behavior action space, its initial version never considered RL as a training paradigm for such generators. As a response, a recent study blended the frameworks of experience-driven PCG and PCG via reinforcement learning, namely ED(PCG)RL; EDRL in short \cite{shu2021experience} focuses on the use of RL for the algorithmic creation of content according to a surrogate model of player experience or affect. 

Whilst there exist a variety of studies on the topic of agent emotion and reinforcement learning, literature on using human-annotated emotion as a training signal for learning is limited \cite{moerland2018emotion}. It has been shown that coupling an agent's simulated affect with its action-selection mechanism allows it to find its goal faster and avoid premature convergence to local optima \cite{broekens2007affect}. Similarly, \cite{hasson2011emotions} showed that using affect as a form of social referencing is a simple method for teaching a robot tasks, such as obstacle avoidance and object reaching. Work on intrinsic motivation through the RL paradigm \cite{singh2005intrinsically,singh2010intrinsically} is also highly relevant to our aims. Intrinsic motivation studies by definition, however, ignore human demonstrations, behavioral and importantly affective \cite{jaques2019social}. A number of very recent studies (e.g. \cite{hussenot2020show}) view the intrinsic motivation paradigm from an inverse RL lens through which reward functions are inferred from behavioral demonstrations.  

The work in this paper expands upon the current state of the art by viewing affect modeling as an RL paradigm and explicitly blending agent behavior and affect using a cutting edge RL algorithm for hard exploration problems. The result is a set of agents which are tested in games in this initial study. The game agents trained to behave (i.e. play) optimally, even better than humans, and ``feel" like a human would (via arousal imitation), or a blend of the two approaches with varying degrees of importance.    

\section{Blending Behaviour and Affect}\label{sec:method}

\begin{figure}[!tb]
\centerline{\includegraphics[width=\columnwidth]{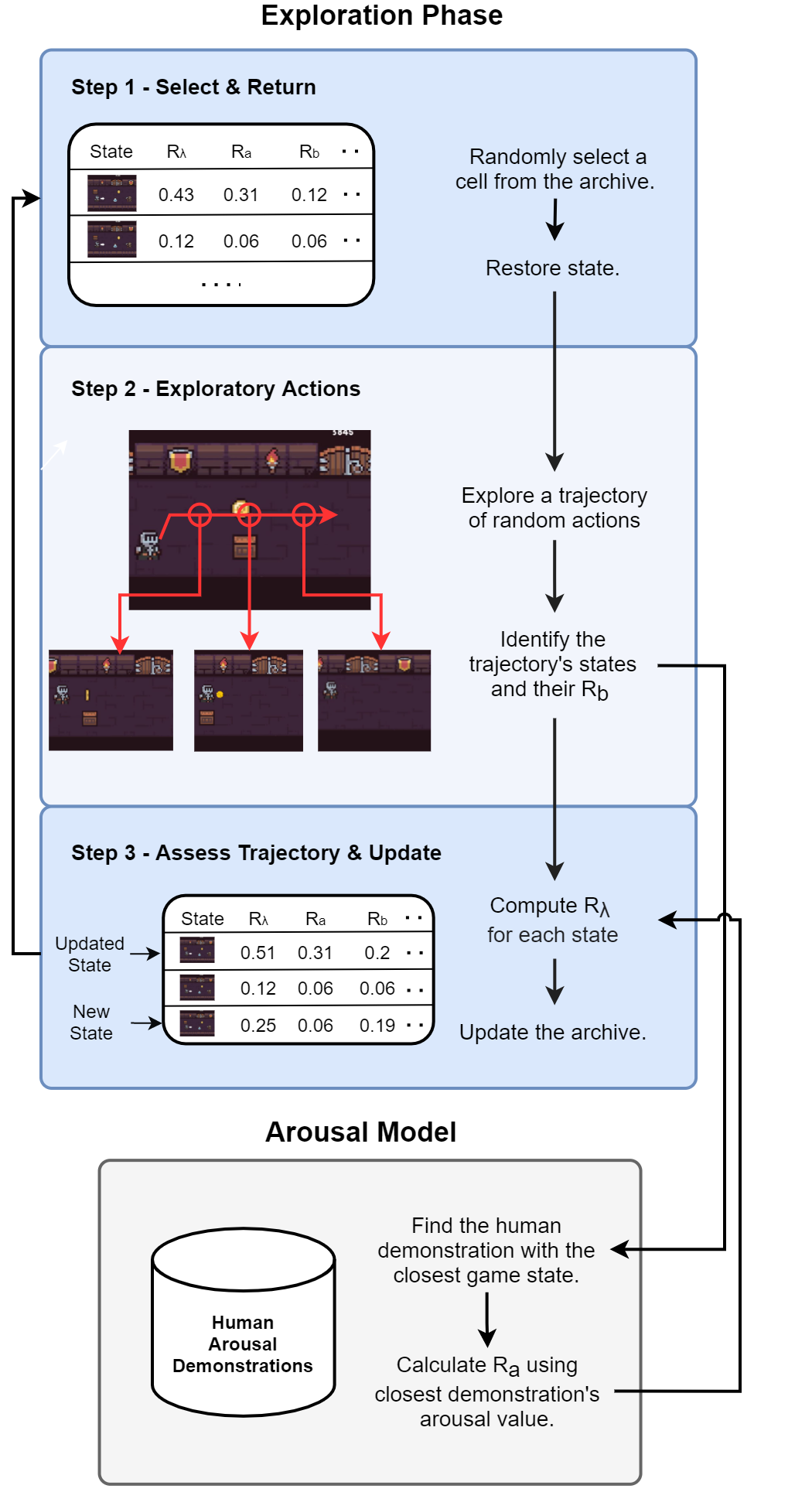}}
\caption{A high-level overview of Go-Explore that blends agent behavior and affect.}
\label{fig:go}
\end{figure}

This paper proposes combining rewards for good behavioral performance with rewards for affect matching in a reinforcement learning agent. We leverage the Go-Explore RL algorithm and describe our implementation in Section \ref{sec:method_goexplore} and how it is enriched with affect information in Sections \ref{sec:method_arousal} and \ref{sec:method_reward}.

\subsection{Go-Explore Implementation}\label{sec:method_goexplore}

The Go-Explore algorithm builds on two phases to create a robust search policy that performs well under a specified reward scheme received from the environment. The first phase is the \emph{exploration} phase, where a deterministic model of the environment is used to explore the search space thoroughly. During exploration an archive of the states encountered so far is used to ensure states are not forgotten, thus preventing the issue of derailment. Each state in the archive also contains the string of actions needed to return to it, addressing the issue of detachment and ensuring that all states can be visited. States are chosen using a selection strategy (e.g. randomly or through the UCB formula \cite{browne2012survey}), after which the algorithm returns to the state as described and begins exploring from there. At its simplest, exploration occurs by taking random actions and updating the cell archive with new states or updating existing ones with better reward values. The move selection strategy can be improved according to the nature of the environment being searched and through the use of expert knowledge.

The result of the exploration phase is a number of high performing trajectories using the deterministic model. If required, the \emph{robustification} phase uses the ``backward algorithm'' \cite{salimans2018learning} to train an agent to perform at the same level (or better) as the trajectories found in exploration, but in a stochastic setting. The backward algorithm is an RL technique used to learn from a given trajectory by decomposing the problem into smaller exploration tasks. It starts by placing the agent near the end of the trajectory and uses an off-the-shelf RL algorithm to train the agent to imitate its last segment. This is repeated several times, moving the starting point further back until the beginning of the trajectory is reached and the agent has been trained on the entire trajectory. To stabilize learning, Go-Explore extends this method to use multiple trajectories which are uniformly sampled at the beginning of each learning episode.

Our implementation of Go-Explore follows the original approach by Ecoffet \textit{et al.} \cite{ecoffet2021first} (see Fig. \ref{fig:go}). An archive of cells stores the game states that have been visited, with each cell representing a unique game state and containing the instructions needed to reach that point in the game. Each cell has an associated reward value, which is used to determine if the cell should be updated in case a similar state with a better score is found. Cells are chosen to explore from randomly, and the actions taken to build trajectories during exploration are also random. Along with the action trajectory to return to its state, each cell also contains trajectories for the state with accompanying cumulative behavioral and affect rewards per trajectory. This implementation of Go-Explore differs from the original version through the inclusion of affect (i.e. arousal in this study) in the reward function. Moreover, in this paper the robustification phase of Go-Explore is not carried out but will be explored in future work.

\subsection{Arousal Model}\label{sec:method_arousal}

A natural question arising when one is asked to blend behavior and affect within a learning process is how the two pieces of information will be considered and fused. An obvious requirement is that the human annotations of affect are time-continuous, thus providing moment-to-moment information about the change of affective states and aligning them with game states stored in playtraces. 

One approach for calculating an affect reward would be to build \emph{a priori} models of affect using supervised learning and use their predicted outcomes \emph{indirectly} as affect-based reward functions. Instead, one could use the affect labels \emph{directly} and build reward functions based on this information. Rather than relying on a trained surrogate model of arousal in a given state, our algorithm queries a dataset of human arousal demonstrations to find the arousal value of the human player closest to the current game state. We use the playtraces and their associated arousal traces directly to assess the player's arousal value in that state which, in turn, provides the intended arousal goal at this point in time.

\subsection{Reward Function}\label{sec:method_reward}

The reward function used for this version of Go-Explore consists of two weighted functions for optimizing behavior and imitating human affect respectively. Both components are normalized within the range $[0,1]$ to avoid uneven weighting between the two objectives. In particular the reward function used, $R_{\lambda}$, is as follows:

\begin{equation}
\label{eq:reward_lambda}
R_\lambda = \lambda \cdot R_a  + (1-\lambda) \cdot R_b 
\end{equation}
\noindent where $R_a$ and $R_b$ are the rewards associated with affect and behavior, respectively, and $\lambda$ is a weighting parameter that blends the two rewards. Formally, the reward associated to affect (i.e. arousal in this paper) is computed as follows:

\begin{equation}
\label{eq:reward_affect}
R_a= \frac{1}{n}\sum^{n}_{i=0} \left(1 - |h(i)-a(i)|\right)
\end{equation}
\noindent where $i$ is a playtrace and affect annotation observation within a time-window; $n$ is the number of observations made so far in this trajectory; $h(i)$ is the agent's estimated arousal value in its current game state; $a(i)$ is the arousal goal at this point in the game. In this paper, we derive $h(i)$ and $a(i)$ directly from human playtraces and their accompanied affect annotations (see Section \ref{sec:method_arousal}). Specifically we calculate $a(i)$ by first creating a \emph{mean arousal trace} averaging all players' arousal values in the same timestamp: this creates a moment-to-moment arousal trace that captures the consensus of players (regardless of actual game context). $a(i)$ is then calculated by finding the arousal value of this mean arousal trace for that time window $i$. On the other hand, $h(i)$ is based on the agent's current game state, finding the annotated arousal value of a human playtrace at any timestamp which has an accompanying game state closest to the agent's game state.

$R_a$ minimizes the absolute difference between the arousal value of a human player in a similar game state as the agent, and the mean annotated arousal value at this time window $i$. Since this difference is averaged across the number of observations made so far, it encourages trajectories with high imitation accuracy across the whole arousal trace generated. 

The reward for the agent's behavior ($R_b$) depends on the game; in this paper we assume that the total score accumulated throughout the game is a sufficient reward for optimal behavior. This assumes that the environment follows arcade game tropes which are played for high-score, as is the case in our case study described in Section \ref{sec:casestudy}. In more complex games, or in games without an explicit score, the reward signal must be designed on an ad-hoc basis such as the reward function used in the original implementation of Go-Explore \cite{ecoffet2021first}.

According to Eq.~\eqref{eq:reward_lambda}, if $\lambda=0$ the reward function trains the agent to only maximise its score (i.e. optimize its behavior) and ignore its associated arousal trace. On the other hand, if $\lambda=1$ the agent is trained to imitate human arousal and ignore its behavior. 

\section{Case Study: Endless Runner}\label{sec:casestudy}

The proposed vision of blending arousal and performance rewards is tested in the ``Endless Runner'' game (hereafter \emph{Endless}). Endless is a platformer game built using the Unity Engine and featured in the AGAIN dataset \cite{melhart2021affect}. The game was chosen for its simple mechanics and objective, and for its accompanying dataset of 112 annotated human play sessions that can be easily used for the arousal model.

\begin{figure}
\centerline{\includegraphics[width=\columnwidth]{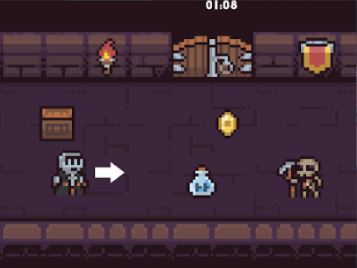}}
\caption{Endless Runner Game Layout}
\label{fig:endless}
\end{figure}

\subsection{Game Description}\label{sec:casestudy_game}

In Endless, the player controls an avatar that constantly moves towards the right and must avoid or destroy obstacles that spawn in their path. The platform consists of two lanes (top/bottom) and the player's only controls is switching lanes by moving up or down (via keyboard input) and/or using a melee attack described below. Game objects are placed on one of two lanes (upper/lower) and are spawned at random intervals. Game objects include items that the player may collide with to improve their score (coins) or alter their movement speed (potions). Other game objects are obstacles (which include immobile enemies); the player must use their melee attack when in close proximity to the obstacle in order to clear it. Colliding with an obstacle results in a 10 point score penalty, destroys any nearby game objects on the screen, and resets the player's speed to the default value. Every 3 seconds the player is passively awarded a point to their game score on top of any bonus points they may receive for collecting coins. Every 10 seconds the speed of the player increases by a fixed amount, increasing the difficulty of the game. In theory, the game can be played for as long as the player wants. During data collection for the AGAIN dataset, an Endless session ended after exactly two minutes and the player has infinite lives. We follow the same duration in all experiments in this paper in order to leverage players' affect annotations and compare the agents' performance with human play.

\subsection{Go-Explore for Endless Runner}\label{sec:casestudy_goexplore}

The game was converted into a deterministic environment to be compatible with the exploration phase of Go-Explore. The sequence of objects to be spawned and their spawn times was fixed to ensure the same sequence of game states are observed when replaying trajectories. Moreover, the game could start from any saved snapshot (i.e. any visited game state). This minimizes the time spent returning to a new cell’s state and allows the algorithm to focus on exploration, an approach central to the Go-Explore paradigm \cite{ecoffet2021first}. To decide which cell the game state should be assigned to, the game state is mapped as an 8-parameter vector describing the player's current lane (two binary values, one per lane), and which game objects are on each lane at specific distance bands (near, mid-distance, and far). The possible values for these bands are empty, item, or obstacle, and in case items and obstacles exist in the same band, it is treated as an obstacle band. The reward for optimal behavior ($R_b$) in Endless is the player's total score after an action is taken. This value is normalized between 0 and 1 with respect to the optimal score achievable in the play session. The optimal in-game score is calculated by summing two components.  The first is the total amount of points awarded to the player passively over time for not dying during the game. The second is the maximum amount of bonus points achievable by picking up every coin in the deterministic environment.

\subsection{Experimental Protocol}\label{sec:casestudy_experiment}

Reported results per method are averaged across five independent runs of the Go-Explore algorithm. Each run consists of the exploration phase of Go-Explore (there is no robustification phase in this first experiment), and the agent returns and explores 4,000 times before selecting the best trajectory and saving it. The agent explores a maximum of 20 actions before choosing a new state to explore from. The actions taken during exploration are chosen at random among the 6 possible options (move up or down, move up or down and attack, no action and attack). The new state to explore from is chosen at random among those already discovered: the reward of the state in the archive, or the number of times it has been visited is not considered. The best trajectories are saved and can be used for the robustification phase of Go-Explore in future work.

The $\lambda$ parameter of Eq.~\ref{eq:reward_lambda} was varied to observe the relationship between learning to play the game optimally and learning to imitate human annotated arousal. Table \ref{tab1} shows the five values used for the $\lambda$ parameter, ranging from 0 to 1 in increments of 0.25. Recall that at $\lambda=0$ and $\lambda=1$ the agent tries to learn to solely behave optimally or to solely ``feel" like a human respectively. 
As a baseline, an experiment with an agent that performed random actions was carried out and results are averaged from 5 independent runs. To estimate this random agent's arousal levels, a trace was generated based on the game states visited using the same approach as in the Go-Explore experiments.

The results were compared to the average performance seen by humans in the dataset for both behavior and arousal reward functions. All results given are the average observed across the 5 runs of Go-Explore, paired with the 95\% confidence interval.

\subsection{Results}\label{sec:casestudy_results}

\begin{table}
\caption{Results for Endless averaged from 5 runs and including the 95\% confidence intervals.}
\begin{center}
\begin{tabular}{|c|c|c|c|}
\hline
\textbf{Experiment}&\multicolumn{3}{|c|}{\textbf{Performance Measures}} \\
\cline{2-4} 
\textbf{Setup} & $R_b$ & $R_a$ & $R_\lambda$ \\
\hline
$R_{0.0}$ & 0.79 ($\pm$0.0474) & 0.72 ($\pm$0.0126) & 0.79 ($\pm$0.0474) \\
\hline
$R_{0.25}$ & 0.73 ($\pm$0.0818) & 0.73 ($\pm$0.0181) & 0.73 ($\pm$0.0569) \\
\hline
$R_{0.5}$ & 0.74 ($\pm$0.0741) & 0.74 ($\pm$0.0145) & 0.74 ($\pm$0.0311) \\
\hline
$R_{0.75}$ & 0.69 ($\pm$0.0658) & 0.76 ($\pm$0.0147) & 0.74 ($\pm$0.0082) \\
\hline
$R_{1.0}$ & 0.25 ($\pm$0.1335) & 0.79 ($\pm$0.0056) & 0.79 ($\pm$0.0056) \\
\hline
\hline
Random & 0.03 ($\pm$0.1012) & 0.75 ($\pm$0.0074) & N/A \\
\hline
Human & 0.70 ($\pm$0.0467) & 0.77 ($\pm$0.0131) & N/A  \\
\hline
\end{tabular}
\label{tab1}
\end{center}
\end{table}

\begin{figure}
\centering
\subfloat[Cumulative Behavior Reward ($R_b$) ]{\includegraphics[width=0.76\columnwidth]{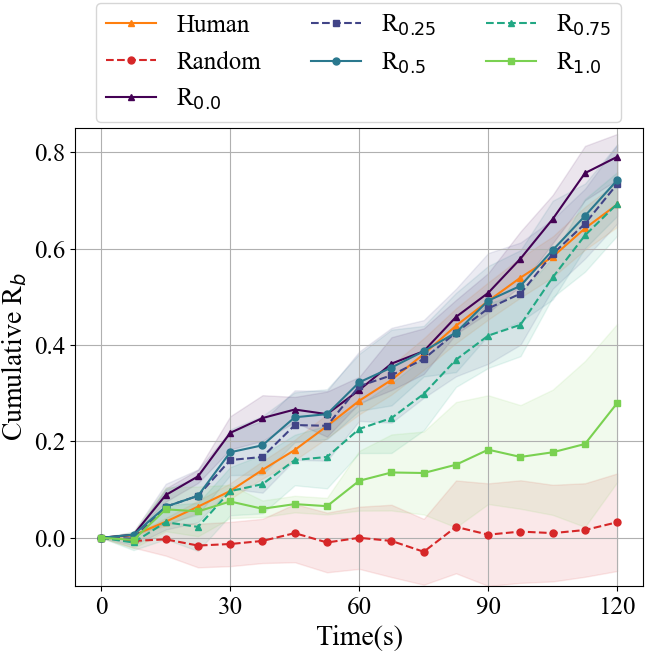}\label{fig:cumulative_behavior}}
\vfill
\subfloat[Cumulative Arousal Reward ($R_a$) ]{\includegraphics[width=0.76\columnwidth]{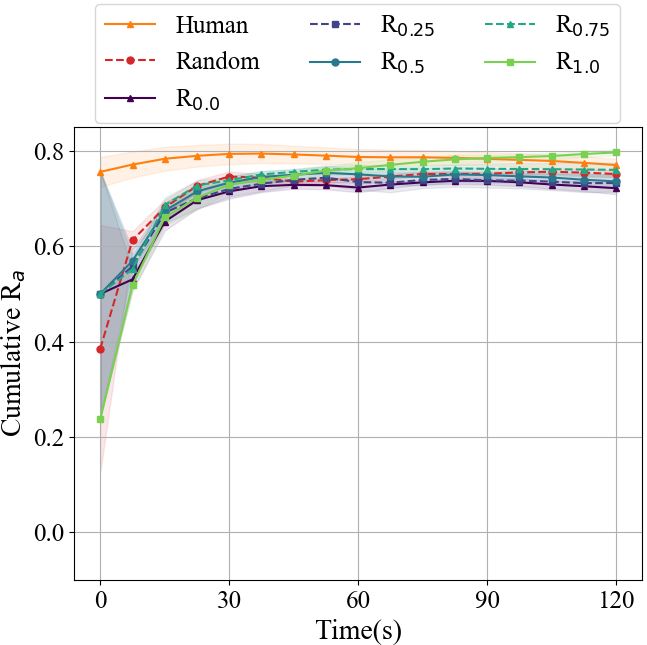}\label{fig:cumulative_arousal}}
\vfill
\subfloat[Cumulative Total Reward ($R_{\lambda}$) ]{\includegraphics[width=0.76\columnwidth]{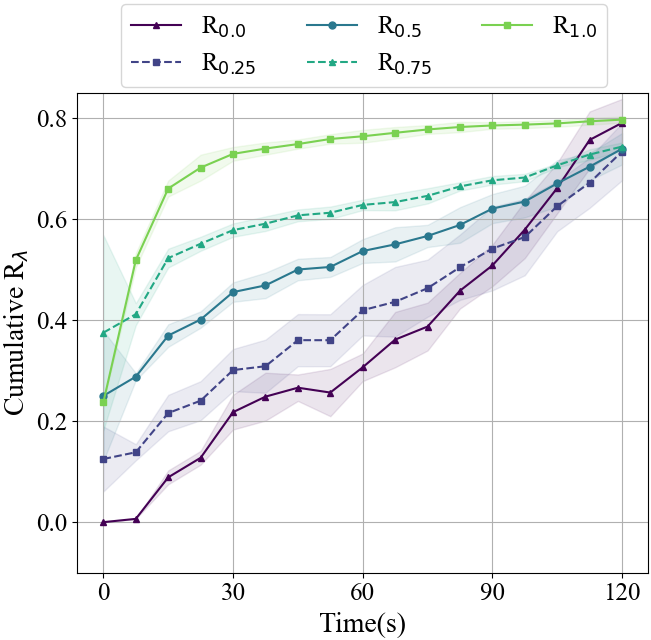}\label{fig:cumulative_total}}
\caption{Cumulative rewards averaged from 5 independent runs. Shaded areas denote the 95\% confidence interval.}
\label{fig:cumulative_all}
\end{figure}

Table \ref{tab1} shows the final values observed for the cumulative behavior ($R_b$) and arousal ($R_a$) components, as well as the overall reward function ($R_\lambda$) for each experiment.
Note that the baseline agent and human entries are not included in the $R_\lambda$ column as they were not trained using Go-Explore. Figure \ref{fig:cumulative_total} illustrates how the agents' overall cumulative reward fluctuates over time for each Go-Explore configuration. Note that due to different $\lambda$ values, the $R_{\lambda}$ values across experiments are not comparable but the differences in how it fluctuates over time provides insight into the behavior of the algorithm. It is clear that agents with higher priority assigned to arousal imitation tend to converge to their maximum value quicker due to the nature of the arousal reward function. Since at $R_{0.0}$ the total reward amounts to a normalized measure of the agent's in-game score, it is not surprising that high scores are only attainable at late points in the game. Instead, states that match the mean arousal trace seem to be easily discovered even early in the game.

Looking at the results for the behavioral component (i.e. the total game score normalized to the absolute best possible score), the random agent shows the worst performance as one would expect when playing most games. While the exploration phase of Go-Explore relies on a random sequence of actions, the discovery of interim states (cells) to explore from and the optimization of these states based on $R_{\lambda}$ clearly leads to a more efficient playstyle than random. For $R_{1.0}$, the agent still manages to produce a better score than the random agent but remains significantly lower than the average human player. Random and $R_{1.0}$ also display a wider confidence interval compared to the rest of the experiments, pointing to an inconsistent behavior. When the behavior component is introduced with a small weight (e.g. $R_{0.75}$), the score immediately matches that of the average human demonstration. As $\lambda$ is lowered to zero, the agent's score improves and surpasses human levels of performance. Figure \ref{fig:cumulative_behavior} illustrates how the agents' cumulative behavior reward changes over time for each configuration. As noted above, the cumulative behavior reward is very time-dependent by design (players reach higher scores the longer they play) but clearly the random agent (and $R_{1.0}$ to a degree) tends to lose score by hitting obstacles which seems to perfectly offset passive score gains.

The results for the arousal component tell a similar story to the results for behavior, with the exception of the random agent. Unsurprisingly, the arousal score increases as $\lambda$ increases from 0 to 1. What is surprising however is the arousal scores attained by the random agent, which seem to be almost at the same levels as the human trace and is only significantly surpassed  by $R_{1.0}$. The potential reasons for this are discussed in section \ref{sec:discussion}. Figure \ref{fig:cumulative_arousal} illustrates how the agents' cumulative arousal reward changes over time for each configuration. It is evident that unlike $R_b$ which is tied to the game score, it is easy to attain high values in $R_b$ early on, and it is also easy to maintain the same levels throughout the game even when performing random actions.

\section{Discussion}\label{sec:discussion}

This paper envisions how affect modeling and expression can be realised thought the RL paradigm. In particular,  we investigate how arousal traces can be used as human demonstrations that train a gameplaying agent to learn how to feel like a human. In the simple testbed of Endless Runner, the large number of annotated playtraces allowed us to match an agent's game state to a human player's game state and use the player's annotated arousal level directly. Results indicate that, as expected, updating the cells of Go-Explore based on the agent's in-game performance (a normalized version of the game score) leads to optimal behavior policies that surpass the average human scores. Combining this performance-based reward with an arousal-based reward that aimed to mimic human annotations resulted in a minor drop of performance which, nevertheless, remains human-competitive. Evidently, using this arousal-based policy alone was detrimental to gameplay performance and points to some limitations of the current way that $R_a$ of Eq.~\ref{eq:reward_affect} is calculated.

The fact that for all agents, including the random action baseline, the cumulative arousal reward swiftly reached high values points to a task that is overly easy. It seems that deriving a policy only based on $R_a$ does not motivate the agent to explore many different states, although the number of updates or new cells encountered in Go-Explore has not been studied sufficiently to verify this hypothesis. Moreover, it should be noted that the human annotations of arousal were processed in an unbounded, ordinal fashion and normalized after the fact. While most players follow a similar pattern of increasing arousal as the game goes on, using the numerical difference between one human's arousal value (closest to the agent's game state) and the mean could reintroduce subjectivity biases due to the normalization applied. Designing another reward function for arousal that better matches the ordinal nature of affect \cite{yannakakis2017ordinal, yannakakis2018ordinal} would be an important direction for future work. Finally, both performance and affect rewards are measured cumulatively, in part due to the fact that the former is the player's score. Exploring different variants by e.g. averaging either score increase or arousal similarity across a narrower time window is expected to have an effect on agents' performance.

It is also worth noting that the Endless Runner testbed has a low branching factor and a deterministic game state. Therefore each experiment was subjected to a very similar sequence of game states. While the simple game still showed that performance-based optimization via Go-Explore is vastly superior to a random agent, it may have affected the arousal model in unexpected ways. Due to the few visited game states, it is likely that the range of values that could be returned by the arousal model was small, which is a likely cause for the agents' similar arousal accuracy and small confidence intervals across the board. Furthermore, when identifying the closest human for the arousal model, a relatively small subset of sessions are used for computational efficiency which further limits the range of arousal values that could be observed. Changing the approach for deriving an arousal value for a given state with this limitation in mind would help generate more diverse traces and allow the differences in the reward functions to become more pronounced. A more complex game where the agent has more degrees of freedom and more arousing stimuli for the human playtesters will also likely illuminate the strengths and weaknesses of this approach.

This proof of concept opens up several avenues for future work to further explore the relationship between behavior and affect in the context of reinforcement learning. Obvious next steps have been highlighted above in terms of refining the arousal reward function and testing the approach in more complex, more stimulating games. Another direction is testing machine-learned predictors of affective states rather than the direct mapping to the closest human trace performed currently. While surrogate models are often inexact, it may counteract the sparse game states encountered by human players when matching an unseen state. More importantly, incorporating the robustification phase in the Go-Explore algorithm is expected to lead to new insights on the impact of affect-based rewards, especially since the environment will no longer be deterministic and thus many more game states are likely to be visited. Finally, imitating human behavior (as a form of reward function) can reveal interesting new relationships between the human-like behavior and affect and optimal play; such derived policies would likely allow agents to play (near) optimally, whilst attempting to imitate both human behavior and human affect. 

\section{Conclusion}\label{sec:conclusion}

This paper presents a proof of concept implementation of a new reinforcement learning paradigm for affective computing where behavioral and affective goals are interwoven. We leverage the Go-Explore algorithm due to its cutting edge ability to solve hard exploration problems, and we pair it with a set of reward functions that blend optimal behavior with arousal imitation to different degrees. Using the Endless Runner game as a platform to test the implementation, we were able to make use of an extensive dataset of human play sessions and accompanying arousal demonstrations that guided the agent's policy. While this initial study focused on a single, simple game, the next steps of our investigations include the enhancement of the Go-Explore approach to cater for its robustification phase, the introduction of ordinal reward functions, and the extension of the approach to accommodate more complex environments within and beyond games.

\bibliographystyle{IEEEbib}
\bibliography{blending_behavior}

\end{document}